%% file: main.tex
\documentclass[preprint,12pt]{elsarticle}

\usepackage{amssymb}
\usepackage{amsmath}
\usepackage{multirow}
\usepackage{booktabs}
\usepackage{colortbl}
\usepackage{xcolor}
\usepackage{url}
\usepackage{float}
\usepackage{algorithm,algorithmic}
\newcommand{\ie}{\emph{i.e.}}
\newcommand{\eg}{\emph{e.g.}}
\newcommand{\darkgreen}[1]{\textcolor[rgb]{0.00,0.70,0.00}{#1}}
\newcommand{\darkblue}[1]{\textcolor[rgb]{0.00,0.00,0.70}{#1}}

\journal{Nuclear Physics B}

\begin{document}

\begin{frontmatter}

\title{SFA: Scan, Focus, and Amplify toward Guidance-aware Answering for Video TextVQA}

\author{Haibin He$^{1}$, Qihuang Zhong$^{1}$, Juhua Liu$^{1\dagger}$, Bo Du$^{1}$, Peng Wang$^{2}$, \\
Jing Zhang$^{1\dagger}$\\
\{haibinhe, zhongqihuang, liujuhua, dubo\}@whu.edu.cn, jingzhang.cv@gmail.com, p.wang@mmu.ac.uk\\
${\dagger}$ Corresponding author
}

\affiliation{organization={$^{1}$School of Computer Science, National Engineering Research Center for Multimedia Software, Institute of Artificial Intelligence, and Hubei Key Laboratory of Multimedia and Network Communication Engineering, Wuhan University},
            city={Wuhan},
            state={Hubei},
            country={China}}

\affiliation{organization={$^{2}$Department of Computing and Mathematics, Manchester Metropolitan University},
            city={Manchester},
            country={UK}}

\input{section/0_abstract}

\end{frontmatter}

\input{section/1_introduction}
\input{section/2_relatedwork}
\input{section/3_method}
\input{section/4_experiments}
\input{section/5_limitation}
\input{section/6_conclusion}
\input{section/7_acknowledgements}



\bibliographystyle{unsrtnat}
\bibliography{ref}

\end{document}

%% file: section/0_abstract.tex
\begin{abstract}
Video text-based visual question answering (Video TextVQA) task aims to answer questions about videos by leveraging the visual text appearing within the videos.
This task poses significant challenges, requiring models to accurately perceive and comprehend scene text that varies in scale, orientation, and clarity across frames, while effectively integrating temporal and semantic context to generate precise answers. Moreover, the model must identify question-relevant textual cues and filter out redundant or irrelevant information to ensure answering is guided by the most relevant and informative cues.
To address these challenges, we propose SFA, a training-free framework and the first Video-LLM-based method tailored for Video TextVQA, motivated by the human process of answering questions. By adaptively scanning video frames, selectively focusing on key regions, and directly amplifying them, SFA effectively guides the Video-LLM’s attention toward essential cues, enabling it to generate more accurate answers. SFA achieves new state-of-the-art results across several public Video TextVQA datasets and surpasses previous methods by a substantial margin, demonstrating its effectiveness and generalizability. 
The source code is publicly available at \url{https://github.com/Hxyz-123/SFA}.
\end{abstract}


%% file: section/1_introduction.tex
\section{Introduction}
\label{secInt}

Visual text plays a pivotal role in scene understanding
, as it not only provides supplementary semantic information~\cite{ye2023dptext, ye2023deepsolo} but also facilitates a deeper understanding of the underlying context, narratives, and relationships within a scene~\cite{wu2022multimodal, zeng2023beyond}. Such information is of paramount value in various practical video-related applications, including text-based content manipulation~\citep{lyu2023fetnet, liu2025textdiff, lu2025artglyphdiffuser}, autonomous driving~\citep{zhang2021character,zablocki2022explainability}, and video retrieval~\citep{dai2025text, wu2025large}. In the context of this, researchers have proposed the Video TextVQA task along with corresponding benchmark datasets~\citep{zhao2022towards, tom2023reading} to foster research into video scene text understanding. However, compared to general video question answering (VideoQA) tasks, Video TextVQA presents greater challenges. It required not only comprehension of the overall video content (content comprehension) but also precise location and recognition of key visual text within the video to generate correct answers (text comprehension). This dual-level comprehension significantly increases the complexity of the task.

\begin{figure*}[t]
  \centering
    \includegraphics[width=1.0\textwidth]{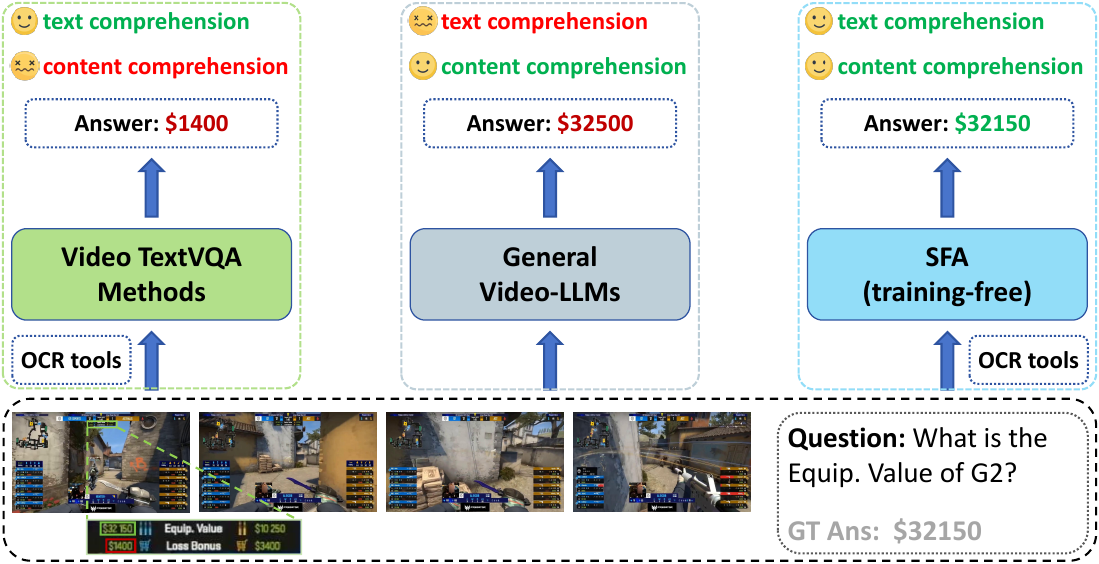}
    \caption{\textbf{An example to illustrate the limitations of both existing Video TextVQA methods and Video-LLMs.} Video TextVQA methods exhibit strong video text perception (text comprehension) capabilities but often misinterpret video content (content comprehension), leading them to select incorrect textual cues when generating answers. Conversely, Video-LLMs demonstrate robust comprehension of video content but limited sensitivity to video text, causing frequent recognition errors and inaccurate responses, particularly when dealing with extreme small text. In contrast, our method achieves dual-level comprehension of both video text and video content, thereby enabling more accurate answers.}
\label{fig:1}
\end{figure*}

In recent years, researchers have proposed several Video TextVQA methods~\citep{zhao2022towards, zhang2025track, zhang2025gather}, following a two-stage paradigm. These methods first leverage OCR systems to extract text-related multimodal features (including textual content, visual information, and positional data), and then employ a Transformer-based answer generator to produce the final responses. For example, T5-ViteVQA~\citep{zhao2022towards} uses five Transformer models and auxiliary tools to encode OCR, video content and question text features, which are subsequently fed into an answer generation module. TEA~\citep{zhang2025track} and GAT~\citep{zhang2025gather} adopt similar strategies. Despite achieving incremental progress in the Video TextVQA field, these methods remain limited by the scale of training data and model parameters, resulting in \textbf{insufficient comprehension of video content} and \textbf{inadequate modeling of visual–textual relationships} (as shown in Fig.~\ref{fig:1}), which in turn leads to low answer accuracy and limited generalization.

Meanwhile, with the expansion of training data and continuous improvements in training strategies, general Video-LLMs~\citep{wang2025internvideo2, bai2025qwen2} have achieved significant breakthroughs in video understanding. However, when directly applied to the Video TextVQA task, their performance remains constrained. We attribute this to the following reasons: 1) Scene text in videos is often \textbf{not the primary subject} (which is typically humans, animals, vehicles, or other objects), causing the general Video-LLMs to easily overlook key textual regions relevant to the question; 2) When multiple scene texts appear in a video, general Video-LLMs may \textbf{attend to focus on incorrect text regions}, thereby affecting the accuracy of the final answer; 3) Due to \textbf{the motion blur, text distortion and the small size of text regions}, general Video-LLMs are prone to recognition errors, introducing noisy information.
In contrast, humans typically adopt a more strategic approach when performing Video TextVQA: we first scan the video to locate potential regions containing key information, then focus on these regions to filter out redundant content, and finally amplify them to obtain the clearer textual details before answering the question. 

Inspired by this, we introduce \textbf{SFA} (\textbf{S}can, \textbf{F}ocus, and \textbf{A}mplify), the first training-free Video-LLM-based method tailored for Video TextVQA. SFA guides the model to selectively attend to essential regions and generate more accurate answers, effectively bridging the gap between general Video-LLMs and the challenges of Video TextVQA. It proceeds in three sequential steps. First, a VTS model~\citep{he2024gomatching, he2025gomatching++} provides the locations of video text in the sampled frames, and an adaptive windowing mechanism is applied to selectively scan these frames based on the detected text positions. Second, a scoring model assesses the relevance of each window to the given question, retaining only the most pertinent window per frame to focus on critical information regions. Finally, the selected windows are amplified to their original frame size to obtain clearer visual details, which are subsequently fed into the Video-LLMs for answer generation. This process emulates the human strategy for answering video text-based questions, providing effective guidance for the model and substantially enhancing its performance. Across multiple public datasets, SFA achieves accuracy significantly surpassing previous methods, with a maximum improvement of 45.80\%, demonstrating both its effectiveness and generalizability.

In summary, the contribution of this paper is threefold:
\begin{itemize}
\item We identify the limitations of existing Video TextVQA methods and Video-LLMs, and introduce \textbf{SFA}, the first training-free Video-LLM-based method tailored for the Video TextVQA task, which integrates visual text perception with video content comprehension to enable more accurate answering.

\item To effectively guide the model’s attention toward key textual regions, we instantiate a three-step \textit{Scan–Focus–Amplify} strategy, inspired by the human question-answering process: it first adaptively scans video frames to identify candidate areas, then filters and selects the most relevant text regions, and finally amplifies these regions to enhance textual clarity and improve answer accuracy.

\item Extensive experiments demonstrate the effectiveness of SFA, which attains new state-of-the-art performance across multiple public datasets and markedly outperforms prior Video TextVQA methods.

\end{itemize}

The remainder of this paper is organized as follows. Section~\ref{secReW} reviews the related work. Section~\ref{secMed} presents the details of the proposed method. Section~\ref{secExp} reports and analyzes the experimental results on publicly available benchmark. Finally, Sections~\ref{secLit} and~\ref{secCon} discuss the limitations of our method and conclude the paper, respectively.

%% file: section/2_relatedwork.tex
\section{Related Work}
\label{secReW}

\subsection{VideoQA}
\label{rel_VidQA}

The development of VideoQA techniques has been largely driven by progress in vision and text representation models. Early neural video models~\citep{xu2017video, li2020hero, li2022invariant, xiao2022video} predominantly adopted a CNN–RNN paradigm, in which videos and texts are encoded by convolutional neural network (\eg, ResNet) and a recurrent neural network (\eg, LSTM), respectively. With the rapid advancement of Transformer in both vision and language domains, a growing number of researchers have developed Transformer-based VideoQA methods~\citep{li2020hero, yang2021just, wang2023all, wang2022git, lei2023revealing} that have achieved remarkable progress. These approaches typically employ a Vision Transformer (\eg, ViT) and a language tokenizer to extract visual and textual tokens, which are then concatenated and fed into a Transformer-based answer generator (\eg, BERT) to produce the final responses. Despite being trained in a multimodal manner, these approaches primarily rely on question text as the sole textual input, neglecting the crucial role of visual text information.

\subsection{Video TextVQA}
\label{rel_ViTQA}

In order to more effectively tackle the Video TextVQA task, researchers have endeavored to develop specialized architectures that integrate OCR tools to extract and utilize visual text features from video frames as additional multimodal inputs for answering. Specifically, T5-ViteVQA~\citep{zhao2022towards} employs multiple Transformer-based models to extract features from OCR results, video content, and question text, which are subsequently fused and fed into an answer generation module to produce the final responses. TEA~\citep{zhang2025track} enhances performance by recovering spatio-temporal relationships among video and emphasizing question-related information through an aggregation module, while incorporating a more sophisticated OCR-based visual feature extractor~\citep{ye2024hi} to refine text representation. In the most recent research, GAT~\citep{zhang2025gather} further advances performance by directly leveraging a powerful video text spotting (VTS) model~\citep{he2024gomatching, he2025gomatching++} to extract visual video text features, integrating a context-aggregated instance gathering module to construct unified textual representations, and utilizing an instance-focused trajectory tracing module to capture spatio-temporal relationships among text instances for accurate answer inference. However, constrained by the scale of training data and the capacity of model parameters, these methods exhibit limited video comprehension capabilities, resulting in low accuracy.

\subsection{Video-LLMs}
\label{rel_VidLLM}

Recent progress in large language mdels (LLMs)~\citep{ cai2024internlm2, team2024qwen2, dubey2024llama} has significantly facilitated the development of Video-LLMs~\citep{lin2024video, cheng2024videollama, liu2025nvila, wang2024qwen2, wang2025internvideo2, bai2025qwen2}, where LLMs serve as foundational backbones to strengthen video understanding capabilities. For example, VideoLLaMA2~\citep{cheng2024videollama} employs CLIP~\citep{radford2021learning} and Qwen2-Instruct~\citep{team2024qwen2}, proposing a Spatial-Temporal Convolution (STC) connector to enhance spatial–temporal modeling. NVILA uses SigLip~\citep{zhai2023sigmoid} and Qwen2~\citep{bai2025qwen2}, and improves VILA~\citep{lin2024vila} by a scale-then-compress strategy that increases spatial and temporal resolution while compressing visual tokens. InternVideo2.5~\citep{wang2025internvideo2}, built upon InternViT~\citep{chen2024expanding} and InternLM2.5~\citep{cai2024internlm2}, is designed to advance fine-grained visual perception and long-range temporal reasoning by incorporating long and rich context (LRC) modeling. Qwen2.5-VL~\citep{bai2025qwen2} introduces dynamic-resolution processing, absolute time encoding, and native dynamic-resolution ViT to enable precise object localization, structured document parsing, and long-video comprehension, thereby enabling efficient multimodal reasoning grounded in Qwen2.5~\citep{team2024qwen2}. Although existing Video-LLMs possess strong video understanding capabilities, they tend to overlook visual text present within videos, resulting in limited performance on Video TextVQA tasks. To enhance the text perception capability of Video-LLM, we propose \textbf{SFA}, a three-step framework designed to guide Video-LLM to attend to critical visual text regions, thereby improving answer accuracy.

%% file: section/3_method.tex
\begin{figure*}[t]
  \centering
    \includegraphics[width=1.0\textwidth]{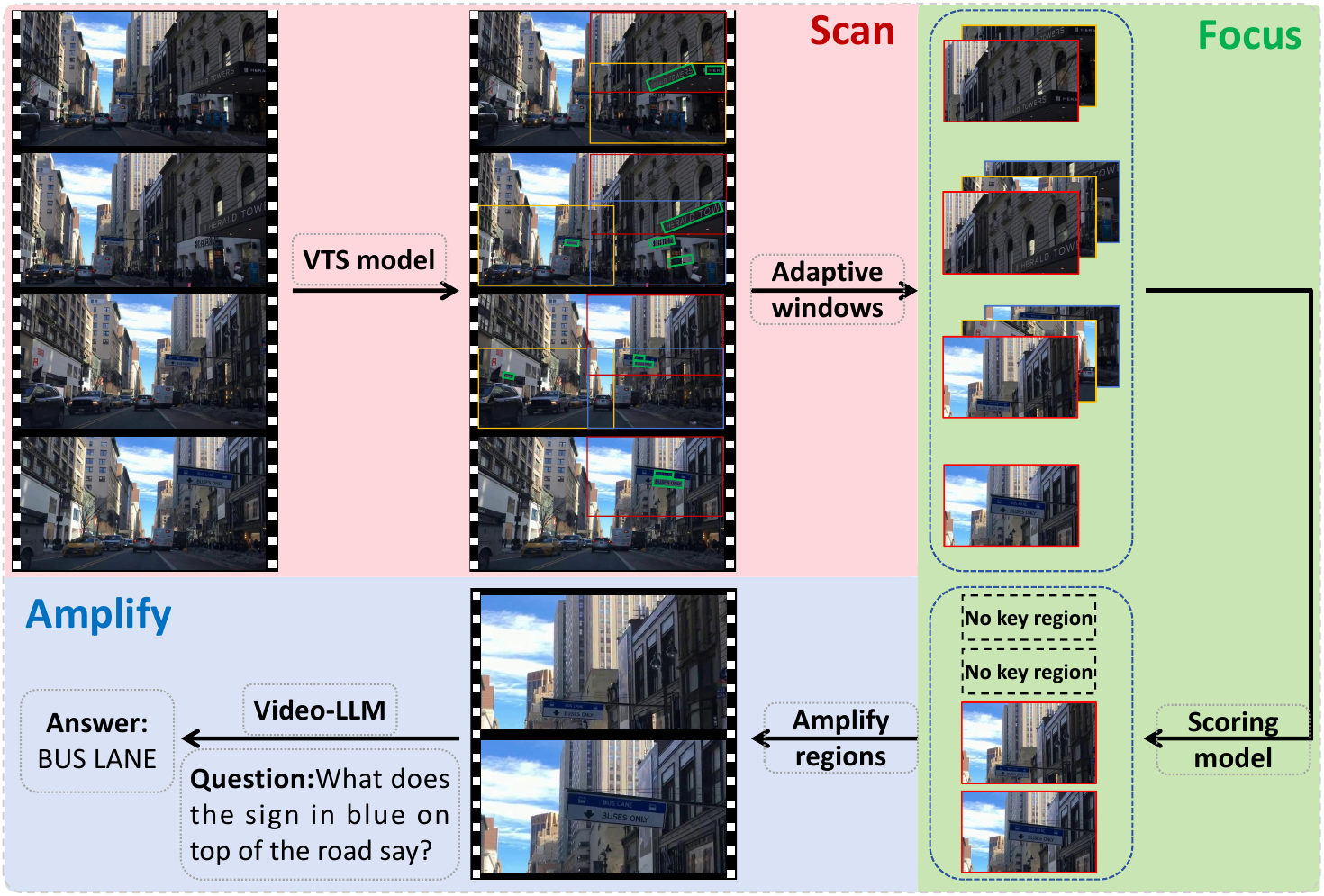}
    \caption{\textbf{The pipeline of SFA.} In the \textbf{Scan} stage, candidate regions containing video text are identified using a well-trained VTS model. An adaptive windowing mechanism is designed to prevent fixed-size windows from fragmenting text, thereby avoiding potential semantic incompleteness and inconsistency. During the \textbf{Focus} stage, a scoring model evaluates the importance of each region and retains at most one most key region per frame. Finally, in the \textbf{Amplify} stage, the selected key regions are restored to the original frame size and fed into the Video-LLM to produce the final answer.}
\label{fig:2}
\end{figure*}

\section{Method}
\label{secMed}

\subsection{Overview}

The overall pipeline of our proposed SFA is depicted in Fig.~\ref{fig:2}, which is designed to guide Video-LLMs toward key textual regions in videos, thereby enhancing their ability to reason over scene text. It contains three pivotal steps: \textbf{Scan}, \textbf{Focus} and \textbf{Amplify}. Specifically, the \textbf{Scan} stage identifies candidate regions containing video text through a well-trained VTS model with an adaptive windowing mechanism to maintain semantic integrity. The \textbf{Focus} stage employs a scoring model to evaluate the importance of these regions and retain the most relevant one per frame. Finally, the \textbf{Amplify} stage resizes the selected region to its original frame size and delivers it to the Video-LLM for answer generation. Notably, SFA is a training-free method and represents the first Video-LLM-based framework tailored for the Video TextVQA task.

\subsection{Scan}
\label{med_scan}
Given a $T$ frames video $V = \{F_1, F_2, ... ,F_T\}$, we first leverage a well-trained textline-level VTS model~\citep{he2024gomatching, he2025gomatching++} $\theta_v$ to detect and localize text lines within each frame. Frames without any detected text are discarded, yielding a text-aware subset $V_{Text} = \{F^1_{Text}, F^2_{Text}, ... ,F^t_{Text}\}$, where $t$ denotes the number of remaining frames containing at least one text line. Subsequently, an adaptive windowing mechanism is applied to scan each remaining frame. Specifically, each frame is partitioned into up to four subregions—top-left, top-right, bottom-left, and bottom-right—according to a predefined scaling ratio $\alpha$ $(0.5 \leqslant \alpha \leqslant 1.0) $. The ratio determines the initial window size as $(w \times \alpha, h \times \alpha)$, where $w$ and $h$ denote the original frame width and height. The window boundaries are then adaptively adjusted based on the location of text lines to ensure semantic integrity. Windows without textual content are further removed, resulting in a refined set of candidate regions containing visual text cues. It can be denoted as $W = \{W_1, W_2, ..., W_t\}$, where each $W_i = \{W_{i1}, ...\}$ represents the set of windows in frame $i$, and $W_{ij}$ means the $j-th$ window in that frame $(1 \leqslant j \leqslant 4) $.

\begin{figure*}[t]
  \centering
    \includegraphics[width=1.0\textwidth]{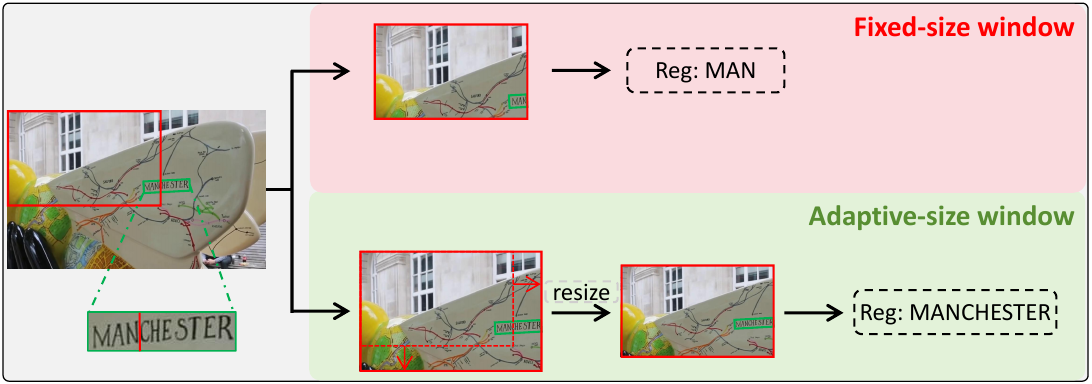}
    \caption{\textbf{Adaptive Windowing Mechanism.} When scanning with fixed-size windows, text lines may be fragmented, potentially altering their semantic meaning (e.g., "MANCHESTER" being recognized as "MAN"). In contrast, adaptive-size windows enable adjustments to their dimensions while preserving the original aspect ratio, ensuring that each text line in the window is fully encompassed and thus maintaining semantic integrity and completeness.}
\label{fig:3}
\end{figure*}

\textbf{Adaptive Windowing Mechanism.} As shown in Fig.~\ref{fig:3}, each scanning window is initially set to a predefined size $(w \times \alpha, h \times \alpha)$. If any text line within the window is partially truncated, the window is adaptively resized while maintaining its original aspect ratio so that it fully encompasses the truncated text line, thereby preserving semantic integrity and completeness. Then, all windows are uniformly resized to the predefined dimensions $(w \times \alpha, h \times \alpha)$ to ensure consistent input scale for subsequent processing.

\subsection{Focus}
\label{med_focus}
To effectively identify key regions while eliminating redundant information, a scoring model is introduced to assess the importance of each remaining window in relation to answer the question $Q$. In particular, a multimodal large language model (MLLM) $\theta_s$ is employed as the scoring model to evaluate the relevance between the visual elements, including objects and visual text, within each window and the question. As formulated in Eq.~\ref{eq:1}, the scoring model produces an relevance score $C_{ij}$ conditioned on the input window $W_{ij}$ and the corresponding question, thereby quantifying the importance of the window.  

\begin{equation}
    C_{ij} = \theta_s(W_{ij}, Q),
    \label{eq:1}
\end{equation}

By applying a predefined threshold $\tau$, windows with low relevance are filtered out, and only the most relevant window in each frame is retained to represent that frame. This process not only reduces redundancy but also guides the model to focus on the most critical regions for accurate reasoning, as formally defined as follows:
\begin{equation}
    W'_{i} = \left\{
    \begin{array}{ll}
        \emptyset, \quad\quad\text{if all } C_{ij} < \tau \\
        W_{ij}, \quad\text{if any } C_{ij} > \tau \text{ and } C_{ij} == \max(C_*)
    \end{array},
    \right.
    \label{eq:2}
\end{equation}

The obtained key region set $W' = \{W'_1, ...\}$ is afterwards passed to the \textbf{Amplify} stage for further processing.

\subsection{Amplify}
\label{med_amplify}
In the \textbf{Amplify} step, all key regions within the set $W'$ are amplified to match the original frame dimensions $(w, h)$, thereby enhancing the clarity of textual details and forming a new video $V' = \{F'_1, ...\}$ that contains only the most relevant information regions. This refined video $V'$ is then paired with the question $Q$ and fed into the Video-LLM $\theta_a$ to generate the final answer $A$. This process can be formulated as:
\begin{equation}
    F'_i = \textit{Resize}(W'_i)  \text{ , for } W'_i \text{ in } W'
    \label{eq:3}
\end{equation}
\begin{equation}
    A = \theta_a (V', Q) \text{ , where } V' = \{F'_1, ...\}
    \label{eq:4}
\end{equation}

The overall pipeline of SFA is illustrated in Algorithm ~\ref{alg:sfa}, providing a clear overview of its key steps and operational flow.

\begin{algorithm}[t!]
  \footnotesize
	\renewcommand{\algorithmicrequire}{\textbf{Input:}}
	\renewcommand{\algorithmicensure}{\textbf{Output:}}
	\caption{SFA: Scan, Focus and Amplify}
	\label{alg1}

	\begin{algorithmic}[1]
            \STATE \textbf{Input:} The input video $V = \{F_1, F_2, ..., F_T\}$, question $Q$, VTS model $\theta_v$, scoring model $\theta_s$, Video-LLM $\theta_a$, window scaling ratio $\alpha$, relevance threshold $\tau$
            \STATE \textbf{Output:} The answer $A$
		\\ $\#\#\#$ Step 1. Scan $\#\#\#$
            \STATE $t = 1$
            \STATE $W = [\ ]$  // candidate region set
		\FOR{Each frame $F_i \in V$}
                \STATE $F_{Text} = \theta_v (F_i)$
			\IF  {$F_{Text}$ contains text} 
                \STATE Obtaining windows contains text $W_{t} = \{W_{t1}, ... \}$ based on adaptive windowing mechanism and $\alpha$
			\STATE $W.\text{append}(W_{t})$
                \STATE $ t = t + 1$
                \ENDIF
		\ENDFOR
            \\ $\#\#\#$ Step 2. Focus $\#\#\#$
              \STATE $W' = [\ ]$ // key region set
		     \FOR{Regions in each frame $W_i \in W$}  
                    \STATE max\_score = 0
                    \STATE $n = 1$
                    \FOR{Each region $W_{ij} \in W_i$}
                        \STATE $C_{ij} = \theta_s(W_{ij}, Q)$ // assess relevance score
                        \IF {$C_{ij} < \tau$}
                            \STATE continue
                        \ELSE
                            \IF {$C_{ij} > \text{max\_score}$}
                                \STATE $W'_n = W_{ij}$
                                \STATE $\text{max\_score} = C_{ij}$
                            \ENDIF
                        \ENDIF
                    \ENDFOR
                    \STATE $W'.\text{append}(W'_n)$
                    \STATE $n = n + 1$
              \ENDFOR
            \\ $\#\#\#$ Step 3. Amplify $\#\#\#$
            \STATE $V' = [\ ]$ // new video $V'$
            \FOR{Each key region $W'_{i} \in W'$}
                \STATE $F'_i = Resize(W'_{i}, (h,w))$ //Amplify key region to the original frame size $(h,w)$
                \STATE $V'.\text{append}(F'_i)$
            \ENDFOR
            \STATE $A = \theta_a(V', Q)$ // use amplified key regions to answer question
	\end{algorithmic}
\label{alg:sfa}
\end{algorithm}

%% file: section/4_experiments.tex
\section{Experiments}
\label{secExp}

\subsection{Experiment Settings}
\subsubsection{Datasets and Evaluation Metrics}
\label{exp_data}
\textbf{M4-ViteVQA}~\citep{zhao2022towards} comprises 8,511 video clips spanning nine categories (\ie, \textit{shopping, traveling, driving, vlog, sport, advertisement, movie, game and talking}) with three kinds of resolutions (\ie, 720p, 1080p and $1176 \times 664$), along with 24,123 question-answer (QA) pairs. The dataset includes 2 tasks (Task1 and Task2) with 3 settings (\textit{Task1Split1, Task2Split2 and Task2}) for robustness evaluation. \textit{Task1Split1} is for regular testing, while \textit{Task1Split2} is for generalization testing, where videos within the same category differ substantially in content. \textit{Task2} is designed for domain adaption, in which the model is trained on seven categories and tested on the remaining two.

\textbf{RoadTextVQA}~\citep{tom2023reading} is a Video TextVQA dataset conducted for the context of driver assistance. It consists of 3,222 driving video clips paired with 10,500 QA instances. All videos are unified to a  resolution of $1280 \times 720$ with a frame rate of 30 frames per second.

\textbf{Evaluation Metrics.} Following the previous Video TextVQA works~\citep{zhao2022towards, tom2023reading, zhang2025track, zhang2025gather}, we adopt two evaluation metrics: \textit{accuracy} and the \textit{average normalized Levenshtein similarity} (ANLS)~\citep{biten2019scene}.

\begin{figure*}[t]
  \centering
    \includegraphics[width=1.0\textwidth]{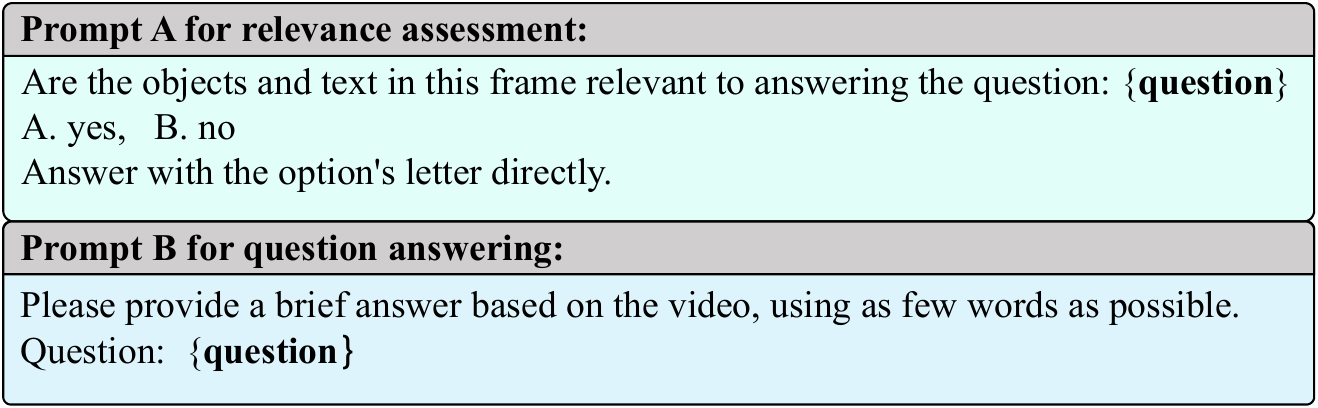}
    \caption{\textbf{The prompts used in SFA.} The upper prompt (A) is designed for relevance assessment in the Focus stage, whereas the lower prompt (B) is employed for question answering in the Amplify stage.}
\label{fig:4}
\end{figure*}

\subsubsection{Implementation Details}
\label{exp_imp}
In the \textbf{Scan} stage,  GoMatching++~\citep{he2025gomatching++} is employed as the VTS spotting model to achieve reliable textline-level video text detection, with the window scaling ratio $\alpha$ set to 0.6. In the \textbf{Focus} stage, Qwen2.5-VL-7B~\citep{bai2025qwen2} serves as the scoring model, where the relevance threshold $\tau$ is set to 0.7. In the \textbf{Amplify} stage, we utilize the same Qwen2.5-VL-7B~\citep{bai2025qwen2} for final answering. The prompts used in the \textbf{Focus} and \textbf{Amplify} stages for assessing the relevance and answering the questions are presented in Fig.4. All the ablation studies are conducted on M4-ViteVQA \textit{Task1Split1} validation set.

\begin{table*}[t]
\centering
\footnotesize
\setlength{\tabcolsep}{2.5pt}
{\begin{tabular}{lc|cccc}
\toprule[1pt]
\multirow{3}{*}{Methods} & \multirow{3}{*}{Year} &\multicolumn{4}{c}{M4-ViteVQA \textit{Task1Split1}} \\

\cline{3-6} & &\multicolumn{2}{c}{Validation} &\multicolumn{2}{c}{Test}\\

\cline{3-6} & &ACC.(\%) &ANLS(\%) &ACC.(\%) &ANLS(\%) \\
\hline

\rowcolor{gray!15} \multicolumn{6}{c}{\textit{Video-language Pretraining Methods}} \\
JustAsk~\citep{yang2021just} &2021 &10.81 &15.40 &10.05 &14.10 \\
All-in-One-B~\citep{wang2023all} &2023 &11.47 &15.30 &10.87 &14.80 \\
\hline

\rowcolor{gray!15} \multicolumn{6}{c}{\textit{Video TextVQA Methods}} \\
T5-ViteVQA~\citep{zhao2022towards} &2022 &23.17 &30.10 &22.17 &29.10 \\
TEA-Base~\citep{zhang2025track} &2025 &34.45 &42.91 &31.70 &40.24 \\
TEA-Large~\citep{zhang2025track} &2025 &37.49 &46.38 &34.78 &43.71 \\
GAT-Base~\citep{zhang2025gather} &2025 &35.31 &44.64 &35.56 &45.21 \\
GAT-Large~\citep{zhang2025gather} &2025 &38.01 &47.53 &38.30 &48.23 \\
\hline

\rowcolor{gray!15} \multicolumn{6}{c}{\textit{Video-LLM Methods}} \\
Video-LLaVA~\citep{lin2024video} &2024 &15.82 &17.77 &15.43 &17.15 \\
VideoLLaMA2~\citep{cheng2024videollama} &2024 &20.04 &21.73 &20.76 &23.55 \\
NVILA~\citep{liu2025nvila} &2025 &37.89 &47.67 &37.73 &47.23 \\
Qwen2-VL-7B~\citep{wang2024qwen2} &2024 &36.77 &46.56 &35.22 &45.84 \\
InternVideo2.5-8B~\citep{zhang2025gather} &2025 &39.83 &48.55 &40.0 &48.79 \\
Qwen2.5-VL-7B~\citep{bai2025qwen2} &2025 &\underline{58.35} &\underline{67.08} &\underline{56.11} &\underline{64.77} \\
\hline

\multirow{2}{*}{\textbf{SFA(ours)}} & \multirow{2}{*}{-} &\textbf{60.98} (\darkblue{$\uparrow$2.63}) &\textbf{68.62} (\darkblue{$\uparrow$1.54}) &\textbf{57.05} (\darkblue{$\uparrow$0.94}) &\textbf{65.44} (\darkblue{$\uparrow$0.67}) \\
& &(\darkgreen{$\uparrow$22.97}) &(\darkgreen{$\uparrow$21.09}) &(\darkgreen{$\uparrow$18.75} ) &(\darkgreen{$\uparrow$17.21}) \\

\hline
\bottomrule[1pt]
\end{tabular}}
\caption{\textbf{Performance comparison of state-of-the-art methods on M4-ViteVQA Task1Split1.} The best and second-best results are marked in \textbf{bold} and \underline{underlined}, respectively. \darkblue{\textbf{Blue}} numbers represent the performance gains of our proposed SFA over the baseline model (Qwen2.5-VL-7B), whereas \darkgreen{\textbf{green}} numbers indicate the improvements achieved compared with the current state-of-the-art Video TextVQA method (GAT).}
\label{tab:1}
\end{table*}

\begin{table*}[t]
\centering
\footnotesize
\setlength{\tabcolsep}{2.5pt}
{\begin{tabular}{lc|cccc}
\toprule[1pt]
\multirow{3}{*}{Methods} & \multirow{3}{*}{Year} &\multicolumn{4}{c}{M4-ViteVQA \textit{Task1Split2}} \\

\cline{3-6} & &\multicolumn{2}{c}{Validation} &\multicolumn{2}{c}{Test} \\

\cline{3-6} & &ACC.(\%) &ANLS(\%) &ACC.(\%) &ANLS(\%) \\
\hline

\rowcolor{gray!15} \multicolumn{6}{c}{\textit{Video-language Pretraining Methods}} \\
JustAsk~\citep{yang2021just} &2021 &7.16 &10.00 &5.47 &8.60  \\
All-in-One-B~\citep{wang2023all} &2023 &6.85 &9.20 &5.66 &7.80  \\

\rowcolor{gray!15} \multicolumn{6}{c}{\textit{Video TextVQA Methods}} \\
T5-ViteVQA~\citep{zhao2022towards} &2022 &17.59 &23.10 &16.68 &23.80  \\
TEA-Base~\citep{zhang2025track} &2025 &26.66 &36.61 &26.29 &36.00  \\
TEA-Large~\citep{zhang2025track} &2025 &28.27 &36.32 &28.43 &38.13  \\
GAT-Base~\citep{zhang2025gather} &2025 &29.07 &39.26 &29.77 &40.71  \\
GAT-Large~\citep{zhang2025gather} &2025 &31.35 &41.33 &30.90 &41.81  \\
\hline

\rowcolor{gray!15} \multicolumn{6}{c}{\textit{Video-LLM Methods}} \\
Video-LLaVA~\citep{lin2024video} &2024 &13.14 &14.29 &11.19 &12.02  \\
VideoLLaMA2~\citep{cheng2024videollama} &2024 &18.30 &19.63 &18.33 &20.45  \\
NVILA~\citep{liu2025nvila} &2025 &30.25 &40.58 &30.10 &41.52  \\
Qwen2-VL-7B~\citep{wang2024qwen2} &2024 &28.55 &39.34 &27.25 &38.45  \\
InternVideo2.5-8B~\citep{zhang2025gather} &2025 &41.60 &52.49 &38.99 &49.36  \\
Qwen2.5-VL-7B~\citep{bai2025qwen2} &2025 &\underline{54.69} &\underline{63.47} &\underline{50.93} &\underline{61.14}  \\
\hline

\multirow{2}{*}{\textbf{SFA(ours)}} & \multirow{2}{*}{-} &\textbf{57.53} (\darkblue{$\uparrow$2.84}) &\textbf{66.21} (\darkblue{$\uparrow$2.74}) &\textbf{55.02} (\darkblue{$\uparrow$4.09}) &\textbf{64.63} (\darkblue{$\uparrow$3.49})  \\
& &(\darkgreen{$\uparrow$26.18}) &(\darkgreen{$\uparrow$24.88}) &(\darkgreen{$\uparrow$24.12} ) &(\darkgreen{$\uparrow$22.82}) \\

\hline
\bottomrule[1pt]
\end{tabular}}
\caption{\textbf{Performance comparison of state-of-the-art methods on M4-ViteVQA Task1Split2.} The best and second-best results are marked in \textbf{bold} and \underline{underlined}, respectively. \darkblue{\textbf{Blue}} numbers represent the performance gains of our proposed SFA over the baseline model (Qwen2.5-VL-7B), whereas \darkgreen{\textbf{green}} numbers indicate the improvements achieved compared with the current state-of-the-art Video TextVQA method (GAT).}
\label{tab:2}
\end{table*}

\begin{table*}[t]
\centering
\footnotesize
\setlength{\tabcolsep}{2.5pt}
{\begin{tabular}{lc|cccc}
\toprule[1pt]
\multirow{3}{*}{Methods} &\multirow{3}{*}{Year} &\multicolumn{4}{c}{M4-ViteVQA \textit{Task2}} \\

\cline{3-6} & &\multicolumn{2}{c}{Validation} &\multicolumn{2}{c}{Test} \\

\cline{3-6} & &ACC.(\%) &ANLS(\%) &ACC.(\%) &ANLS(\%) \\
\hline

\rowcolor{gray!15} \multicolumn{6}{c}{\textit{Video-language Pretraining Methods}} \\
JustAsk~\citep{yang2021just} &2021 &4.86 &6.70 &3.60 &6.70 \\
All-in-One-B~\citep{wang2023all} &2023 &4.20 &5.00 &3.28 &4.60 \\

\rowcolor{gray!15} \multicolumn{6}{c}{\textit{Video TextVQA Methods}} \\
T5-ViteVQA~\citep{zhao2022towards} &2022 &12.30 &16.10 &9.29 &13.60 \\
TEA-Base~\citep{zhang2025track} &2025 &20.73 &28.18 &17.28 &26.03 \\
TEA-Large~\citep{zhang2025track} &2025 &22.83 &30.21 &18.83 &28.90 \\
GAT-Base~\citep{zhang2025gather} &2025 &21.65 &30.88 &21.65 &29.83 \\
GAT-Large~\citep{zhang2025gather} &2025 &24.54 &33.30 &22.13 &30.75 \\
\hline

\rowcolor{gray!15} \multicolumn{6}{c}{\textit{Video-LLM Methods}} \\
Video-LLaVA~\citep{lin2024video} &2024 &10.89 &13.23 &9.38 &11.80 \\
VideoLLaMA2~\citep{cheng2024videollama} &2024 &19.68 &23.62 &16.54 &21.80 \\
NVILA~\citep{liu2025nvila} &2025 &23.79 &32.89 &22.89 &30.34 \\
Qwen2-VL-7B~\citep{wang2024qwen2} &2024 &22.95 &32.65 &21.23 &28.79 \\
InternVideo2.5-8B~\citep{zhang2025gather} &2025 &48.03 &57.98 &41.36 &51.21 \\
Qwen2.5-VL-7B~\citep{bai2025qwen2} &2025 &\underline{66.40} &\underline{73.28} &\underline{62.98} &\underline{71.00} \\
\hline

\multirow{2}{*}{\textbf{SFA(ours)}} &\multirow{2}{*}{-} &\textbf{70.34} (\darkblue{$\uparrow$3.94}) &\textbf{76.60} (\darkblue{$\uparrow$3.32}) &\textbf{64.46} (\darkblue{$\uparrow$1.48}) &\textbf{71.72} (\darkblue{$\uparrow$0.72}) \\
& &(\darkgreen{$\uparrow$45.80}) &(\darkgreen{$\uparrow$43.30}) &(\darkgreen{$\uparrow$42.33}) &(\darkgreen{$\uparrow$40.97})\\

\hline
\bottomrule[1pt]
\end{tabular}}
\caption{\textbf{Performance comparison of state-of-the-art methods on M4-ViteVQA Task2.} The best and second-best results are marked in \textbf{bold} and \underline{underlined}, respectively. \darkblue{\textbf{Blue}} numbers represent the performance gains of our proposed SFA over the baseline model (Qwen2.5-VL-7B), whereas \darkgreen{\textbf{green}} numbers indicate the improvements achieved compared with the current state-of-the-art Video TextVQA method (GAT).}
\label{tab:3}
\end{table*}

\begin{table*}[t]
\centering
\footnotesize
\setlength{\tabcolsep}{20pt}
{\begin{tabular}{lc|cc}
\toprule[1pt]
\multirow{2}{*}{Methods} &\multirow{3}{*}{Year} &\multicolumn{2}{c}{RoadTextVQA} \\

\cline{3-4} & &ACC.(\%) &ANLS(\%)\\
\hline

\rowcolor{gray!15} \multicolumn{4}{c}{\textit{Video-language Pretraining Methods}} \\

GIT~\citep{wang2022git} &2022 &29.58 &35.23 \\
SINGULARITY~\citep{lei2023revealing} &2023 &24.62 &30.79 \\
\hline

\rowcolor{gray!15} \multicolumn{4}{c}{\textit{Video TextVQA Methods}} \\
TEA-Base~\citep{zhang2025track} &2025 &44.43 &51.69 \\
TEA-Large~\citep{zhang2025track} &2025 &48.14 &54.85 \\
GAT-Base~\citep{zhang2025gather} &2025 &46.54 &53.78 \\
GAT-Large~\citep{zhang2025gather} &2025 &50.23 &58.12 \\
\hline

\rowcolor{gray!15} \multicolumn{4}{c}{\textit{Video-LLM Methods}} \\
Video-LLaVA~\citep{lin2024video} &2024 &30.82 &40.92 \\
VideoLLaMA2~\citep{cheng2024videollama} &2024 &25.11 &36.53 \\
NVILA~\citep{liu2025nvila} &2025 &49.98 &57.22 \\
Qwen2-VL-7B~\citep{wang2024qwen2} &2024 &47.23 &55.34 \\
InternVideo2.5-8B~\citep{zhang2025gather} &2025 &41.96 &49.05 \\
Qwen2.5-VL-7B~\citep{bai2025qwen2} &2025 &\underline{50.33} &\underline{58.38} \\
\hline

\multirow{2}{*}{\textbf{SFA(ours)}} &\multirow{2}{*}{-} &\textbf{61.18} (\darkblue{$\uparrow$10.85}) &\textbf{67.28} (\darkblue{$\uparrow$8.90}) \\
& &(\darkgreen{$\uparrow$10.95}) &(\darkgreen{$\uparrow$9.16}) \\

\hline
\bottomrule[1pt]
\end{tabular}}
\caption{\textbf{Performance comparison of state-of-the-art methods on RoadTextVQA.} The best and second-best results are marked in \textbf{bold} and \underline{underlined}, respectively. \darkblue{\textbf{Blue}} numbers represent the performance gains of our proposed SFA over the baseline model (Qwen2.5-VL-7B), whereas \darkgreen{\textbf{green}} numbers indicate the improvements achieved compared with the current state-of-the-art Video TextVQA method (GAT).}
\label{tab:4}
\end{table*}

\begin{figure*}[t]
  \centering
    \includegraphics[width=1.0\textwidth]{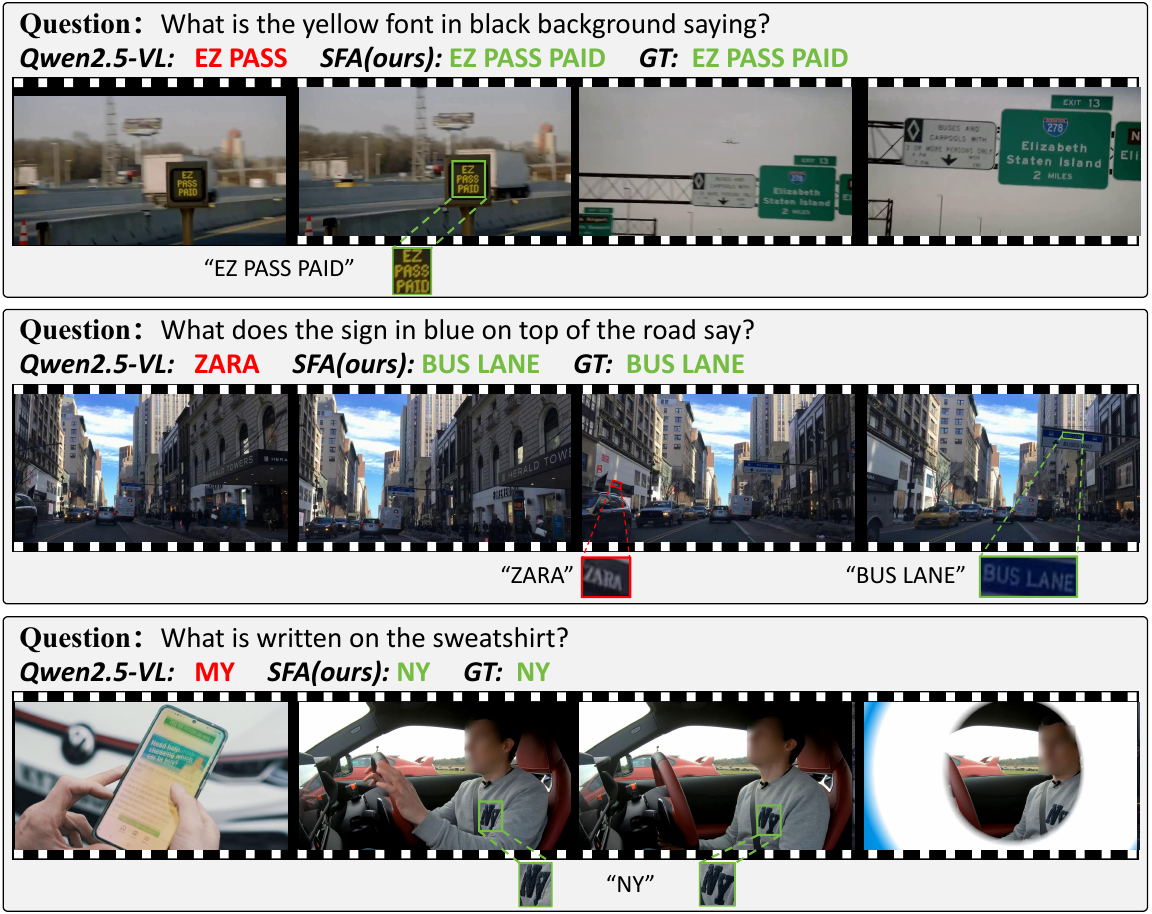}
    \caption{\textbf{Case studies.} Qwen2.5-VL tends to overlook critical text, focus on irrelevant regions, or misrecognize text, whereas the proposed SFA effectively mitigates these issues.}
\label{fig:5}
\end{figure*}

\subsection{Comparison with State-of-the-art Methods}
\label{exp_comp}
Consistant with prior studies~\citep{zhang2025track, zhang2025gather}, we evaluate SFA with serval categories methods: 1) \textbf{Video-language pretraining methods}~\citep{yang2021just, wang2023all, wang2022git, lei2023revealing}; 2) \textbf{Video TextVQA methods}~\citep{zhao2022towards, zhang2025track, zhang2025gather}; and 3) \textbf{Video-LLM methods}~\citep{lin2024video,cheng2024videollama,liu2025nvila,wang2024qwen2,wang2025internvideo2,bai2025qwen2}. Owing to their strong zero-shot capabilities, InternVideo2.5~\citep{wang2025internvideo2} and Qwen2.5-VL~\citep{bai2025qwen2} are evaluted without finetuning on the corresponding datasets. The results are presented in Tab.~\ref{tab:1} to Tab.~\ref{tab:4}.

\textbf{Video-language Pretraining Methods.} Due to their limited ability to read video text and comprehend the video content, video-language pretraining methods demonstrate low performance on Video TextVQA task. Specifically, All-in-One-B~\citep{wang2023all} achieves only 11.47 and 10.87 in accuracy on the M4-ViteVQA Task1Split1 validation and test set, respectively. Moreover, their accuracies on the M4-ViteVQA Task1Split2 and Task2 setting remain below 10. Similarly, GIT~\citep{wang2022git} attains an accuracy of merely 29.58 on the RoadTextVQA validation set. These results collectively highlight the inherent limitations of video-language pretraining methods in addressing the Video TextVQA task.

\textbf{Video TextVQA Methods.} With the incorporation of powerful OCR tools, Video TextVQA methods have achieved notable improvements in identifying scene text within videos, resulting in substantial performance gains over video-language pretraining methods. However, their video understanding capabilities remain limited due to constraints in dataset scale and model parameter capacity, leading to suboptimal performance and generalization. For instance, the current state-of-the-art Video TextVQA method, GAT~\citep{zhang2025gather}, attains accuracies of 38.01, 31.15, and 50.23 on the M4-ViteVQA Task1Split1 validation set, Task1Split2 validation set, and RoadTextVQA validation set, respectively. In contrast, on the cross-domain M4-ViteVQA Task2 setting, GAT achieves only 24.54 and 22.13, revealing its limited generalization ability.

\textbf{Video-LLM Methods.} Early Video-LLM methods~\citep{lin2024video,cheng2024videollama,liu2025nvila,wang2024qwen2} exhibited inferior performance on the Video TextVQA task, primarily due to the insufficient of video training and OCR-related data. However, with the integration of large-scale image-based OCR data, richer video datasets, and improved training strategies, subsequent Video-LLM approaches have achieved significant performance gains, tremendous surpassing existing Video TextVQA methods and demonstrating remarkable zero-shot generalization capabilities. For instance, Qwen2.5-VL~\citep{bai2025qwen2} achieves over 50\% accuracy across all evaluation datasets and reaches 66.40 and 62.98 on the validation and test sets of M4-ViteVQA Task2, respectively, which assess domain adaption ability. Nevertheless, their performance remains bounded by insensitivity to video text, often resulting in incorrect answers caused by misreading or mislocating key textual regions within videos.

\textbf{SFA.} To overcome the limitations in Video-LLM methods, the proposed SFA framework integrates a state-of-the-art Video Text Spotting (VTS) model with a three-stage pipeline — Scan, Focus, and Amplify — to explicitly guide Video-LLM in identifying and emphasizing key textual regions. This design markedly enhances the model’s capability to comprehend and reason over text-intensive video content. Compared with the baseline model~\citep{bai2025qwen2}, SFA consistently achieves superior results across all benchmarks, including a remarkable accuracy improvement of 10.85 on the RoadTextVQA validation set, which predominantly features small and blurred text in driving scenarios. This improvement demonstrates that SFA effectively bridges the performance gap caused by the model’s insensitivity to video text, highlighting its capability to robustly capture challenging visual-text cues in complex, dynamic environments. Furthermore, SFA outperforms the current state-of-the-art Video TextVQA method~\citep{zhang2025gather} by 45.80 and 42.33 on the validation and test sets of M4-ViteVQA Task2, respectively. As shown in Fig.~\ref{fig:5}, Qwen2.5-VL exhibits limitations in accurately perceiving and interpreting video textual information, often overlooking essential text, misfocusing on irrelevant visual text regions, and producing erroneous text recognition. Our SFA effectively mitigates these shortcomings through enhancing video text-aware ability, enabling the model to concentrate on critical text regions and consequently achieve higher answer accuracy. These results provide strong empirical evidence for the effectiveness, robustness, and generalizability of the proposed framework.

\subsection{Ablation Studies}
\label{exp_abl}

\begin{table*}[t]
\centering
\small
\setlength{\tabcolsep}{6pt}
{\begin{tabular}{lcc|cc}
\toprule[1pt]
\multirow{2}{*}{\#} &\multirow{2}{*}{Window Size} &\multirow{2}{*}{Scoring Model} &\multicolumn{2}{c}{M4-ViteVQA \textit{Task1Split1} Validation}\\

\cline{4-5} & & &ACC.(\%) &ANLS(\%) \\
\hline

1 & Fixed &MLLM &57.03 &66.53 \\
2 & Adaptive &LLM &57.58 &66.05 \\
3 & Adaptive &MLLM &\textbf{60.98} &\textbf{68.68} \\
\hline

\bottomrule[1pt]
\end{tabular}}
\caption{\textbf{Effectiveness of the adaptive windowing mechanism and MLLM-based scoring model.} ``Fixed'' and ``Adaptive'' represent the use of a fixed or adaptive window size in the \textit{Scan} stage, respectively. ``LLM'' and ``MLLM'' indicate whether a large language model or a multimodal large language model is used as the scoring model in the \textit{Focus} stage. The best and second-best results are marked in \textbf{bold}.}
\label{tab:5}
\end{table*}

\subsubsection{Effectiveness of the Adaptive Windowing Mechanism and MLLM-based Scoring Model}

Comparing the first and third rows in Tab.~\ref{tab:5}, it can be observed that employing an adaptive window during the Scan stage yields a 3.95\% improvement in accuracy. This improvement arises because the adaptive window dynamically adjusts its size to encompass complete text lines, thereby preserving the semantic integrity and contextual coherence of textual information. In contrast, using a fixed window tends to fragment text lines, which may disrupt the continuity of linguistic meaning and hinder the model’s ability to correctly associate fragmented text segments across frames. This fragmentation often results in incomplete textual understanding, ultimately reducing the accuracy of downstream answering.

Moreover, as shown by comparing the second and third rows, adopting a MLLM as the scoring model leads to a 3.4\% performance gain over using a LLM. The reason lies in the inherent limitation of LLMs, which rely solely on textual content to assess relevance and thus fail to capture vital visual cues. Such text-only evaluation can lead to ambiguities in scenarios where visual grounding is essential. For example, when asked, “What is the price of the eggplant in the video?”, an LLM without access to visual information cannot accurately link the correct price to the eggplant among multiple textual instances of prices.By contrast, the MLLM integrates both visual semantics and textual context when performing relevance estimation, enabling it to jointly reason over visual-textual associations. This multimodal integration ensures that the scoring process is more contextually grounded, reliable, and robust, particularly in questions where visual cues are tightly coupled with textual semantics.

\begin{table*}[t]
\centering
\footnotesize
\setlength{\tabcolsep}{2pt}
{\begin{tabular}{l|c|c|c|cc}
\toprule[1pt]
\multirow{2}{*}{\#} &\multirow{2}{*}{Base Model} &\multirow{2}{*}{Scoring Model} &\multirow{2}{*}{Fps Sampling} &\multicolumn{2}{c}{M4-ViteVQA \textit{Task1Split1} Validation}\\

\cline{5-6} & & & &ACC.(\%) &ANLS(\%) \\
\hline

\rowcolor{gray!15} 1 &Qwen2.5-VL-7B &- &1 &58.35 &67.08  \\
2 &Qwen2.5-VL-7B & Qwen2.5-VL-3B & 1 &59.97 (\darkgreen{$\uparrow$1.62}) &67.91 (\darkgreen{$\uparrow$0.83})  \\
3 &Qwen2.5-VL-7B & Qwen2.5-VL-7B &1 &60.98 (\darkgreen{$\uparrow$2.63}) &68.62 (\darkgreen{$\uparrow$1.54})  \\
\rowcolor{gray!15}4 &Qwen2.5-VL-7B & - & 2 &62.25 &70.0  \\
5 &Qwen2.5-VL-7B &Qwen2.5-VL-7B &2 &63.42 (\darkgreen{$\uparrow$1.17}) &70.77 (\darkgreen{$\uparrow$0.77})  \\
\rowcolor{gray!15}6 &Qwen2.5-VL-32B & - & 1 &55.30 &65.03  \\
7 &Qwen2.5-VL-32B & Qwen2.5-VL-7B &1 &56.27 (\darkgreen{$\uparrow$0.97}) &65.57 (\darkgreen{$\uparrow$0.54})  \\
\hline

\bottomrule[1pt]
\end{tabular}}
\caption{\textbf{Generality analysis results.} The base models used for answering (baseline) are highlighted with a gray background. The experiments evaluate the generalization ability of the proposed SFA method under different settings, including varying the parameter scales of the scoring model, adopting different sampling strategies, and employing base models with diverse parameter sizes. The \darkgreen{\textbf{green}} values in parentheses denote the performance gains relative to the corresponding base model.}
\label{tab:6}
\end{table*}

\subsubsection{Generality Analysis}
We conducted experiments under various configurations to verify the generalizability of the proposed SFA method, including different sizes of scoring models, distinct frame sampling strategies, and base models with varying parameter scales. The experiment results are shown in Tab.~\ref{tab:6}. When using Qwen2.5-VL-3B as the scoring model, SFA achieves an improvement of 1.62 in accuracy compared with directly answering using the base model. With a larger scoring model, Qwen2.5-VL-7B, the performance further increases by 1.01. Under a frame sampling rate of 2 FPS (as shown in the fifth row of Tab.~\ref{tab:6}), SFA also yields an accuracy gain of 1.17 over the corresponding baseline. Moreover, when employing Qwen2.5-VL-32B as the base model, SFA improves accuracy by 0.97 when using Qwen2.5-VL-7B as scoring models. These results collectively demonstrate the robust generalizability and scalability of the proposed SFA.

\begin{figure*}[t]
  \centering
    \includegraphics[width=1.0\textwidth]{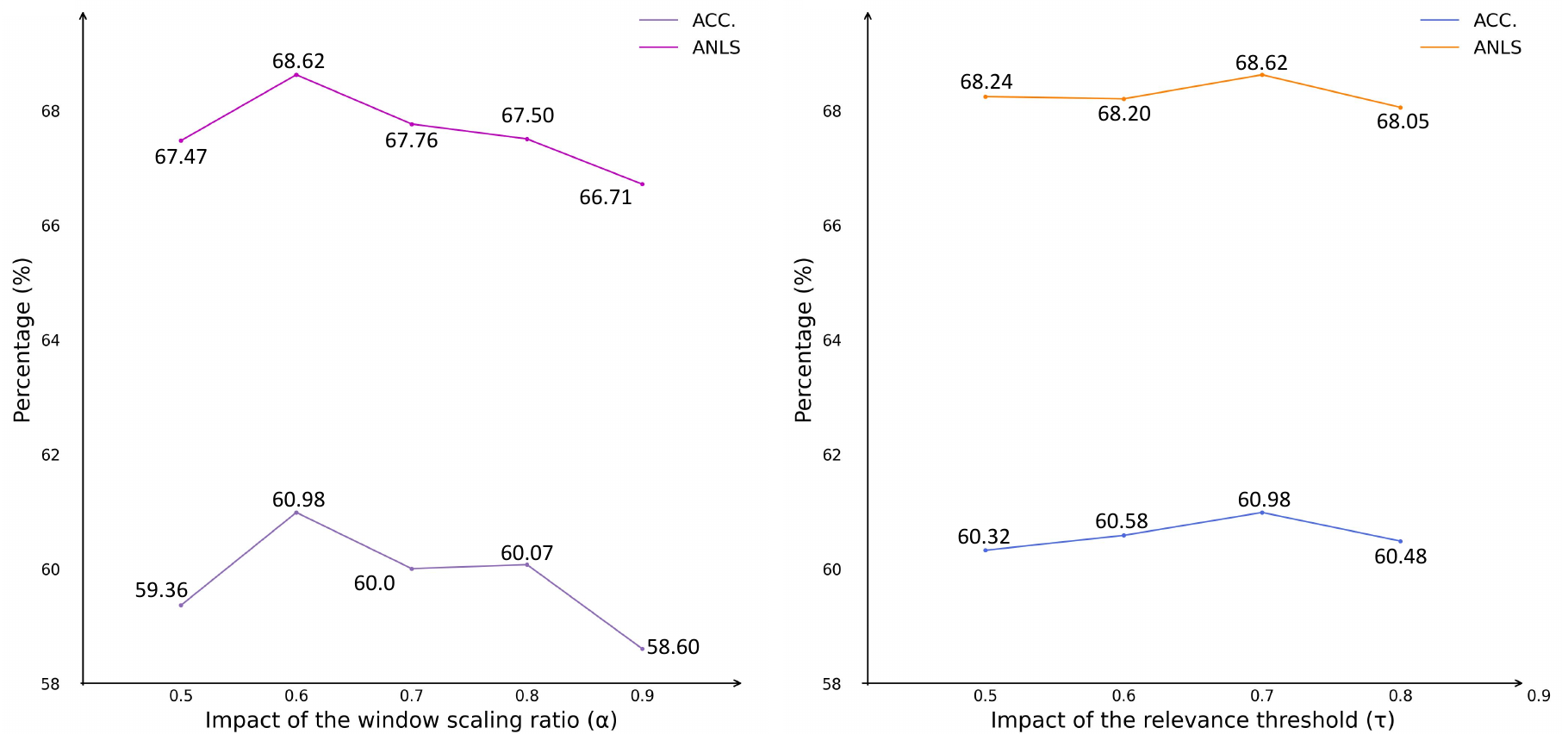}
    \caption{\textbf{Impact of the window scaling ratio and relevance threshold.} }
\label{fig:6}
\end{figure*}

\subsubsection{Impact of the Window Scaling Ratio and Relevance Threshold}
As illustrated in Fig.~\ref{fig:6}, the optimal window scaling ratio is determined to be 0.6, while the relevance threshold achieves the best balance at 0.7. It can be observed that beyond the window scaling ratio of 0.6, performance gradually declines as the window scaling ratio increases. This degradation arises because larger window sizes result in a smaller magnification factor during the \textit{Amplify} stage, leading to reduced clarity of small key text regions. Regarding the relevance threshold, setting it too low introduces excessive irrelevant visual information, thereby increasing noise and reducing the answer accuracy. Conversely, an excessively high threshold raises the risk of mistakenly discarding essential key regions that contain text crucial for answering the question. Hence, a moderate threshold of 0.7 achieves a favorable trade-off between preserving critical textual cues and filtering redundant content.

Furthermore, the stability of performance around these optimal values indicates the robustness of our SFA framework. This suggests that the method is not overly sensitive to parameter variations, which is desirable for deployment across diverse video domains with varying text density and motion dynamics.

%% file: section/5_limitation.tex
\section{Limitation}
\label{secLit}
Although our proposed SFA method significantly enhances the performance of Video-LLMs on the Video TextVQA task, its overall capability remains bounded by the inherent limitations of the current VTS models and the base Video-LLMs. In the future, as Video-LLMs continue to evolve with improved capabilities in video text identification, spatial-temporal reasoning, and fine-grained visual grounding, the reliance on external VTS models may gradually diminish. In such a scenario, SFA could be further simplified into a fully end-to-end framework, seamlessly integrating text perception and reasoning within a unified model.

%% file: section/6_conclusion.tex
\section{Conclusion}
\label{secCon}
In this paper, we identify the limitations of existing methods and introduce SFA, the first training-free Video-LLM-based framework tailored for the Video TextVQA task. Imitating the human question-answering process, SFA employs a three-stage ``Scan–Focus–Amplify” strategy to explicitly guide Video-LLMs toward question-relevant textual regions within videos, effectively mitigating distractions from irrelevant content and enhancing answer accuracy. Extensive experiments on public Video TextVQA benchmarks validate the effectiveness and generalizability of SFA. We hope this work will foster further research on integrating fine-grained text perception with holistic video content comprehension, paving the way toward more capable and reliable video understanding systems.

%% file: section/7_acknowledgements.tex
\section*{Acknowledgements}
This work was supported in part by the National Key Research and Development Program of China under Grant 2023YFC2705700, in part by the National Natural Science Foundation of China under Grants U23B2048 and 623B2076, in part by the Innovative Research Group Project of Hubei Province under Grant 2024AFA017, and in part by the Science and Technology Major Project of Hubei Province under Grants 2024BAB046 and 2025BCB026. The numerical calculations in this paper have been done on the supercomputing system in the Supercomputing Center of Wuhan University. This work was also supported by WHU-Kingsoft Joint Lab.